\documentclass{article}

\usepackage[preprint]{neurips_2026}

\usepackage[utf8]{inputenc} 
\usepackage[T1]{fontenc}    
\usepackage{hyperref}       
\usepackage{url}            
\usepackage{booktabs}       
\usepackage{amsfonts}       
\usepackage{nicefrac}       
\usepackage{microtype}      
\usepackage{xcolor}         
\usepackage[most]{tcolorbox}
\usepackage{listings}
\usepackage{graphicx}
\usepackage{amsmath}
\usepackage{multirow}
\usepackage{algorithm}       
\usepackage{algorithmicx}    
\usepackage{stfloats}
\usepackage[labelformat=simple]{subcaption}

\newtcolorbox{codebox}[1][]{
  colback=gray!5,
  colframe=black!50,
  boxrule=0.6pt,
  arc=2pt,
  left=18pt,
  right=6pt,
  top=6pt,
  bottom=6pt,
  breakable,
  enhanced,
  title=#1,
  fonttitle=\bfseries
}
\counterwithout{table}{section}

\title{Kernel Foundry: A Diagnosis-driven Evolutionary Kernel Optimizer with Multi-Experts}

\author{
  Zixuan Huang\textsuperscript{1}\textsuperscript{*}, Da Chen\textsuperscript{2}\textsuperscript{*}, Kecheng Huang\textsuperscript{1},  Lihao Yin\textsuperscript{2}, Xing Li\textsuperscript{2},\\ \textbf{Huiling Zhen\textsuperscript{2}, Mingxuan Yuan\textsuperscript{2}, Zili Shao\textsuperscript{1}} \\
  \textsuperscript{1}The Chinese University of Hong Kong\\ \textsuperscript{2}Noah’s Ark Lab, Huawei \\
  Hong Kong\\
  \texttt{\{zxhuang, kchuang21, shao\}@cse.cuhk.edu.hk} \\
  \texttt{chenda.shenzhen@gmail.com}\\
  \texttt{\{yin.lihao, li.xing2, zhenhuiling2, yuan.mingxuan\}@huawei.com} \\
  \thanks{
\textbf{These authors contributed equally to this work.} }
}

\begin{document}

\maketitle





\begin{abstract}
Generating high-performance GPU kernels remains challenging due to the need for both correctness and hardware-aware optimization. 
While large language models (LLMs) show promise in code generation, they often fail to produce kernels that are both correct and efficient.

We propose \textbf{Kernel Foundry}, a diagnosis-driven evolutionary framework for automatic GPU kernel optimization. 
Our method combines expert-guided, retrieval-augmented initialization with a multi-island evolutionary search, where candidate kernels are iteratively refined using structured diagnostic feedback. 
A centralized experience library accumulates reusable optimization knowledge to guide subsequent evolution, while explicit mechanisms prevent cheating behaviors that bypass kernel-level computation.

Experiments on KernelBench show that our method consistently improves both correctness and performance over strong baselines, achieving up 100\% correctness on Level~2.
\end{abstract}

\section{Introduction}

GPUs are central to modern computing, supporting workloads from deep learning to large-scale simulations~\cite{paszke2019pytorch,abadi2016tensorflow,vaswani2017attention}. 
Their performance is largely determined by GPU kernels, whose efficiency depends on parallelism, memory access, and synchronization. 
However, writing kernels that are both correct and high-performance remains challenging even for experts, as small design choices can lead to correctness bugs or significant performance degradation~\cite{li2025autotriton,chen2018tvm}.

Recent advances in large language models (LLMs) offer a promising direction for automating kernel development through code generation~\cite{novikov2025alphaevolve}. 
However, directly applying LLMs to kernel synthesis remains insufficient for achieving both correctness and high performance.

Prior work explores reinforcement learning, hardware-aware optimization, agentic search, and supervised generation. 
RL-based methods such as CUDA-L1~\cite{li2025cuda} and Kevin~\cite{baronio2025kevin} improve correctness via execution feedback but suffer from unstable performance gains. 
Hardware-aware approaches such as SwizzlePerf~\cite{tschand2025swizzleperf} and agentic systems (e.g., The AI CUDA Engineer~\cite{lange2025ai}) can achieve strong performance but often incur high search or verification costs and limited generalization. 
Supervised models such as KernelLLM~\cite{kernelllm2025} achieve high correctness in familiar domains but lack mechanisms for systematic optimization beyond initial generation.

\begin{figure*}
    \centering
    \includegraphics[width=\linewidth]{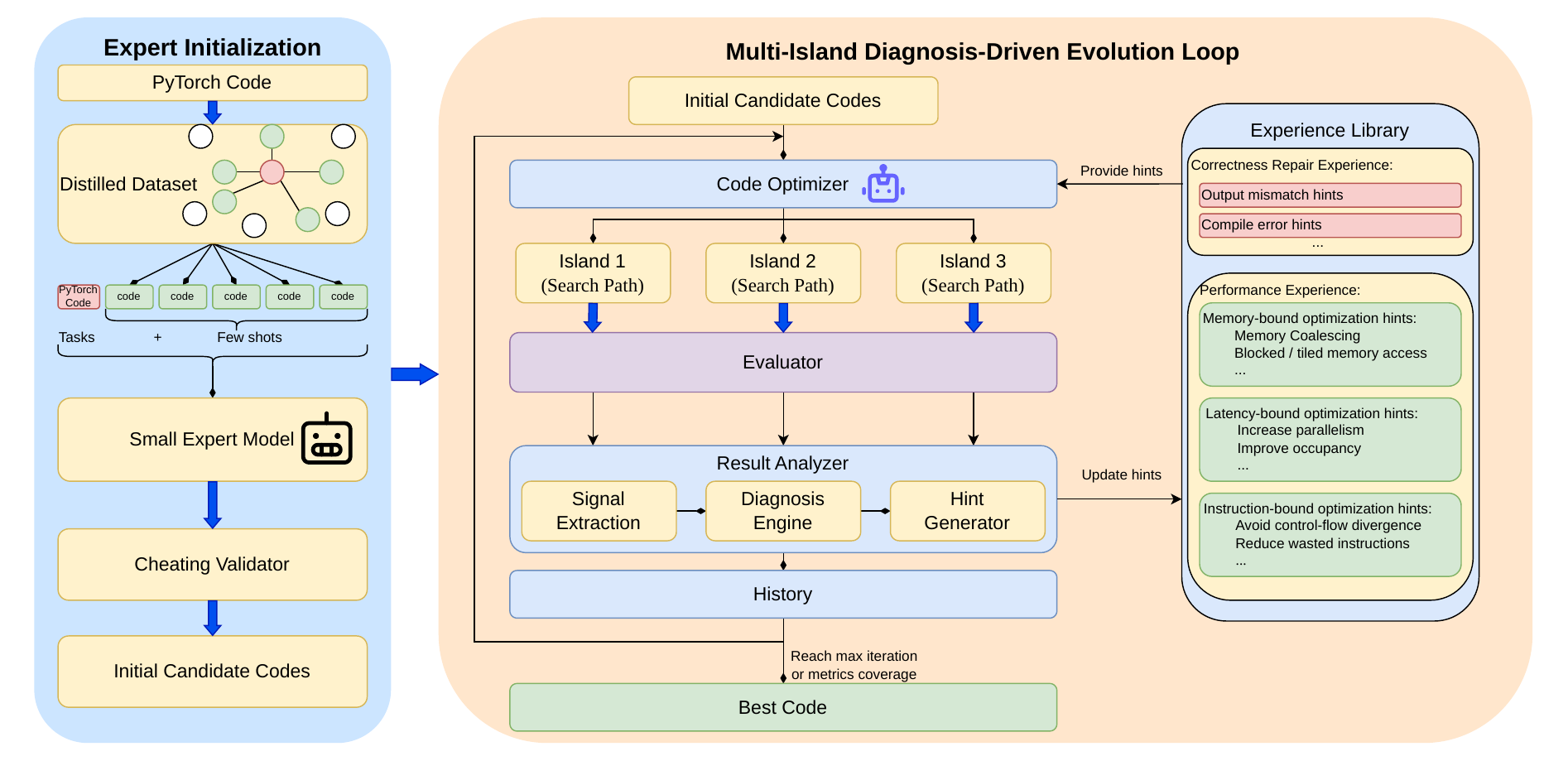}
\caption{Overview of Kernel Foundry. Given a PyTorch operator, the framework initializes candidate Triton kernels using expert-guided, retrieval-augmented priors, and then refines them through a multi-island, diagnosis-driven evolutionary loop. Each island maintains an independent population and optimization trajectory, guided by diagnostic feedback and retrieved hints from a shared experience library. Candidate kernels are iteratively evaluated, diagnosed for errors or performance bottlenecks, and refined until convergence to a high-performance implementation.}
    \label{fig:overview}
    \vspace{-0.6cm}
\end{figure*}

These limitations indicate that achieving both correctness and high performance in automatic GPU kernel
generation requires more than one-shot code synthesis. However, satisfying these requirements is challenging.
The GPU kernel search space is vast, where minor errors in indexing or parallel configuration can easily lead
to compilation or runtime failures, and general-purpose LLMs often lack sufficient GPU- and Triton-specific
knowledge to produce valid kernels.
Even when correctness is achieved, performance optimization over tiling, parallelism, and memory access is
highly non-convex and provides weak optimization signals.
This difficulty is further exacerbated by cheating implementations that exploit evaluation loopholes
(e.g., delegating computation to external libraries), which may appear correct or fast but collapse the
kernel-level optimization space and mislead the search process.

Motivated by these challenges, we propose \emph{Kernel Foundry}, a diagnosis-driven evolutionary framework for
automatic GPU kernel optimization.
As illustrated in Figure~\ref{fig:overview}, given a PyTorch operator, Kernel Foundry initializes candidate
Triton kernels using domain-specialized expert knowledge and retrieval-augmented priors, and then iteratively
refines them through parallel multi-island evolution.
Each candidate is evaluated, diagnosed for correctness issues or dominant performance bottlenecks, and improved
using optimization experience retrieved from a centralized knowledge library.
By filtering cheating behaviors and grounding evolution in structured diagnostic feedback, the
framework progressively converges to genuinely optimized GPU kernels.


To evaluate the effectiveness of our framework, we conduct comprehensive experiments on KernelBench~\cite{ouyang2025kernelbench}, a standardized benchmark suite for GPU kernel generation and optimization. Our framework achieves up to 100\% correctness on level 2 while delivering consistently strong performance, demonstrating that expert-guided initialization and diagnosis-driven evolution can jointly ensure correctness and enable effective performance optimization.

\paragraph{Contributions.}
We summarize the contributions below:
\begin{itemize}
    \item \textbf{Diagnosis-driven evolutionary kernel optimization.}
    We propose a multi-island evolutionary framework for kernel optimization that integrates structured result diagnosis and experience-guided feedback to systematically improve correctness and performance.
    
    \item \textbf{Expert-guided initialization via distillation and retrieval.}
    We distill high-quality expert kernels into a compact corpus and leverage relevance-aware retrieval to guide a domain-specialized expert model toward generating correct initial kernels.
    
    \item \textbf{Experience accumulation for reusable optimization knowledge.}
    We design a structured experience library that organizes optimization knowledge by diagnosed error types and performance bottlenecks, enabling the evolutionary process to continuously refine both kernels and optimization strategies.
    
    \item \textbf{Extensive evaluation on KernelBench.}
    Experiments on KernelBench Level~1 and Level~2 show that our method consistently achieves high correctness and substantial speedups, outperforming strong LLM-based and automated kernel generation baselines.
\end{itemize}

\section{Background}
\subsection{GPU Kernel}

A GPU kernel is a fine-grained function executed in parallel across many threads on a GPU, determining how computation, memory access, and synchronization are mapped to hardware. Although GPUs expose massive parallelism, kernel performance is highly sensitive to design choices such as block size and memory layout, and strongly depends on hardware characteristics including memory hierarchy and scheduling policies. As a result, writing kernels that are both correct and efficient requires substantial expertise, and even minor mistakes can lead to significant performance degradation or subtle correctness issues~\cite{nvidiaCudaProgGuide,fung2007dynamic}. High-level DSLs such as Triton~\cite{tillet2019triton} reduce programming complexity by providing abstractions for memory access and parallelization, but writing highly optimized Triton kernels still requires careful tuning, motivating automated kernel generation and optimization.

\subsection{LLMs for GPU Kernels}

Large language models (LLMs) have demonstrated strong capabilities in program synthesis and code generation~\cite{novikov2025alphaevolve}, motivating their application to GPU kernel programming. Recent studies show that LLMs can generate kernels that compile and implement intended functionality, thereby lowering the barrier to kernel development~\cite{kernelllm2025,li2025autotriton}.

However, generating high-performance GPU kernels remains challenging. Kernel efficiency depends on hardware-specific factors such as memory access patterns, parallel execution behavior, and synchronization, which are weakly represented in generic training data. As a result, existing LLM-based approaches achieve correctness but struggle to deliver consistent performance improvements.

To address these limitations, prior work has explored learning- and search-based optimization methods. Reinforcement learning (RL) approaches, such as CUDA-L1~\cite{li2025cuda}, demonstrate that correctness can be learned from execution feedback, but performance optimization remains difficult due to sparse, noisy, and hardware-dependent signals. AutoTriton~\cite{li2025autotriton} combines supervised learning with RL to refine Triton kernels, yet still relies on unstable runtime feedback.

Evolutionary and agent-based methods further explore kernel variants through mutation and selection~\cite{lange2025ai}, offering greater flexibility but often suffering from inefficiency caused by large search spaces and costly validation. Moreover, coarse-grained performance signals provide limited guidance for identifying optimizations.

Overall, existing LLM-based and learning-driven approaches highlight the potential of automated GPU kernel generation and optimization, but lack mechanisms for structured feedback and reusable optimization knowledge, motivating the diagnosis-driven evolutionary framework proposed in this work.

\section{Methodology}

\subsection{Overview}
\label{sec:overview}
We propose a diagnosis-driven evolutionary framework for automatic GPU kernel generation and optimization.
Given a PyTorch code, the framework synthesizes high-performance Triton kernels by combining expert-guided initialization, multi-island evolutionary search, and structured diagnostic feedback.
Rather than treating kernel generation as a one-shot synthesis problem, our approach formulates it as an iterative optimization process, where candidate kernels are repeatedly evaluated, diagnosed, and refined.

As illustrated in Figure~\ref{fig:overview}, the framework consists of three key components:
(1) expert-guided initialization, which produces a diverse set of valid Triton kernels using domain-specialized models and retrieval-based demonstrations;
(2) multi-island evolutionary optimization, which explores diverse optimization trajectories in parallel with diagnosis-driven feedback; and
(3) an experience library, which accumulates reusable optimization knowledge to guide subsequent evolution.
To ensure that optimization reflects genuine kernel-level improvements, the framework incorporates explicit mechanisms to detect cheating implementations.

\subsection{Kernel Initializer}
\label{sec:init}
\subsubsection{Initializer}

The kernel initializer is not only responsible for providing a correctness-preserving kernel code, but also targets laying a strong foundation for subsequent kernel optimization. To provide a strong and correctness-preserving starting point for evolutionary optimization, our system performs retrieval-augmented initialization. This stage leverages both a distilled corpus of verified Triton kernels and a small expert model specialized in GPU programming. They significantly reduce the difficulty of generating valid kernel seeds.

\textbf{Distillation.}
KernelBook\cite{kernelbook2025} is a dataset consisting of a large number of PyTorch–Triton kernel pairs. To get high-quantity codes, we validate all the codes and only kernels evaluated correctly are retained. These kernels encode essential GPU programming patterns such as index derivation, block tiling, memory coalescing, and parallelism strategies, forming the foundation for retrieval-based initialization.

\textbf{Embedding.}
To enable semantic retrieval, both the distilled Triton kernels and PyTorch queries must be represented in a shared vector space. We adopt GraphCodeBERT~\cite{guo2020graphcodebert} embeddings, which jointly model syntactic structure and data-flow dependencies, critical signals for GPU kernel semantics. In this way, each distilled Triton kernel is converted into a code graph and embedded into a dense vector. Then, these vectors are stored in a Milvus~\cite{wang2021milvus} vector database, enabling efficient search over the expert corpus.

\textbf{Retrieve.}
Given a target PyTorch kernel, we first encode its original Torch code using the same GraphCodeBERT encoder.
A search over the Milvus vector index retrieves the top-5 most semantically similar Triton kernels based on dataflow patterns, tensor access structure, and computational similarity.

These retrieved kernels serve as few-shot demonstrations that expose the expert model to correct memory-access patterns, compatible parallelization strategies and representative Triton idioms for similar workloads. By including these examples in the prompt, the system provides strong structural priors before any evolutionary refinement begins.
\subsubsection{Cheating Validation}

\paragraph{Cheating Behaviors.}
During initialization, models may generate kernels that appear fast but bypass actual computation by exploiting evaluation loopholes. Typical cases include calling external libraries (e.g., PyTorch), defining unused kernels, inserting no-op functions, or omitting core operations and returning constants. Although such implementations may pass superficial checks, they do not reflect genuine kernel optimization and can mislead the search process.

\paragraph{Anti-Cheating Mechanism.}
We enforce anti-cheating at two levels. First, prompt-level constraints require a valid \texttt{@triton.jit} kernel with all computations explicitly implemented in Triton, disallowing external library calls. Second, we apply an LLM-based validator that semantically compares the generated kernel with the reference implementation. It checks for missing computation and verifies genuine Triton execution patterns (e.g., \texttt{tl.load}, \texttt{tl.store}, grid indexing), and outputs a cheating likelihood score. Candidates exceeding a threshold (50\%) are discarded. This mechanism is applied throughout initialization and evolution to ensure that observed speedups arise from real kernel-level optimization.

\subsection{Kernel Optimization}
\label{sec:opti}





\subsubsection{Evolution}
\label{sec:evolution}

\paragraph{Overview.} After initialization, validated kernels enter an evolutionary optimization stage. This stage iteratively refines kernel performance by combining LLM-based mutation, multi-island exploration, and feedback-driven selection. Unlike combining multiple code snippets into a single program, our framework does not merge kernels across islands. Instead, all candidates are evaluated independently, and the final output is selected as the best-performing valid kernel from the union of all island populations. 

\paragraph{Multi-Island Search.} Each evolutionary island maintains an independent \emph{population} of kernel candidates and evolves them over multiple iterations, rather than generating isolated code samples. Unlike naive multi-sample generation from LLMs, where each candidate is produced independently, an island represents a \emph{persistent optimization trajectory} with its own history, selection, and mutation process. Within each island, candidates are iteratively refined based on performance feedback and diagnostic signals, forming a structured evolutionary loop. Different islands are further differentiated by role-specialized system prompts and by retrieving different subsets of optimization hints from the experience library. This leads each island to focus on distinct optimization perspectives, such as operator fusion, memory access optimization, or kernel parameter tuning. As a result, islands explore the search space in parallel but along diverse and structured directions, enabling both deep local refinement within each island and complementary global exploration across islands. 

\paragraph{Iterative Refinement.} Within each island, evolution proceeds in discrete iterations. At each iteration, the language model generates new kernel variants conditioned on the current candidate, historical evolution context, and performance feedback from previous iterations. Generated kernels are compiled and executed on real hardware to measure correctness and runtime. High-performing candidates are preserved as elites and reused as mutation parents, gradually steering the population toward improved kernel implementations. 

\paragraph{Elite Migration.} To balance exploration and exploitation across evolutionary trajectories, we introduce a controlled elite migration mechanism. Each island maintains a local archive of elite kernels discovered along its trajectory. When progress stagnates, elite candidates from other islands are probabilistically migrated and injected as new seeds or mutation parents, allowing effective optimization patterns to propagate while preserving island-level diversity.


\subsubsection{Diagnosis and Experience-Guided Feedback}
\label{sec:diagnosis}
\begin{figure*}
    \centering
    \includegraphics[width=\linewidth]{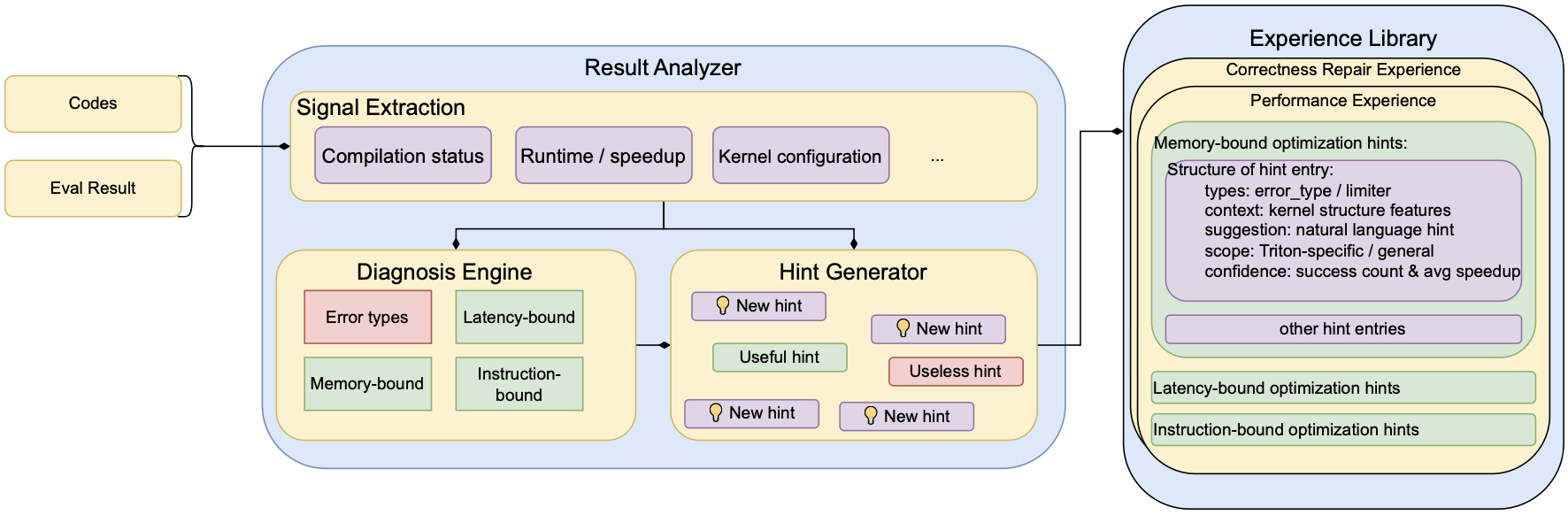}
    \caption{Diagnosis and experience-guided feedback.
Given generated kernels and their evaluation results, the Result Analyzer extracts lightweight execution signals and performs correctness and performance diagnosis to identify error types or dominant performance limiters.
A hint generator assesses the effectiveness of existing hints and distills new optimization hints from consistently improving kernels.
Validated hints are stored in a centralized Experience Library, which accumulates reusable correctness repair and performance optimization knowledge to guide subsequent evolutionary iterations.}
    \label{fig:analyzer}
\vspace{-0.5cm}
\end{figure*}
To guide evolution beyond blind mutation, our framework incorporates a structured diagnosis and feedback mechanism that transforms raw evaluation outcomes into actionable optimization signals.
As illustrated in Figure~\ref{fig:analyzer}, this mechanism is implemented by a Result Analyzer that performs signal extraction, diagnosis, and experience update for each evaluated kernel.

\paragraph{Signal Extraction.}
Given a generated Triton kernel and its execution results, the Result Analyzer first extracts lightweight signals, including compilation status, runtime and speedup, kernel configuration parameters, and execution metadata.
These signals provide a compact yet informative summary of kernel behavior without requiring heavy profiling or intrusive instrumentation.

\paragraph{Diagnosis Engine.}
Based on the extracted signals, the diagnosis engine performs two types of analysis.
For incorrect kernels, it classifies failure modes into distinct error types (e.g., compilation errors or runtime failures), enabling targeted correctness repair.
For correct kernels, it identifies the dominant performance limiter by categorizing kernels as memory-bound, latency-bound, or instruction-bound using runtime statistics and kernel configuration features.
This explicit diagnosis provides an interpretable explanation of why a kernel underperforms and serves as the basis for targeted optimization.

While this classification is approximate and does not rely on detailed hardware profiling, it provides sufficiently informative signals to guide optimization. 
Our goal is not to precisely model hardware behavior, but to provide coarse-grained guidance that steers the search toward promising optimization directions.

\paragraph{Hint Generation and Evaluation.}
Diagnosed results are passed to a hint generator that produces optimization suggestions in natural language.
Hints may target correctness repair or performance improvement, and are associated with structured metadata describing their scope (e.g., Triton-specific or general), applicable context, and expected optimization focus.
Importantly, the framework evaluates the usefulness of existing hints by correlating their usage with observed performance improvements.
Hints that consistently lead to speedup gains are reinforced, while ineffective or misleading hints are down-weighted or discarded.

\subsection{Domain Knowledge Collection}
\label{sec:domain}

The Experience Library serves as a centralized repository of structured optimization knowledge that enables experience-guided evolutionary search. 
It is both expert-initialized and continuously refined during evolution. 
Additional examples are provided in Appendix.

\paragraph{Expert-Initialized Library}
At initialization, the library is constructed by distilling expert knowledge from established GPU optimization principles, primarily derived from NVIDIA documentation. 
These include best practices such as memory coalescing, parallelism tuning, latency hiding, and instruction-level optimization. 
We abstract these principles into reusable optimization hints that provide strong priors for early-stage evolution.

\paragraph{Online Experience Refinement}
During evolution, the library is incrementally updated. 
When a kernel achieves stable performance improvements, its optimization trajectory is analyzed to extract key changes, such as modifications to parallelization, memory access patterns, or kernel configurations. 
These changes are distilled into reusable hints, validated, and stored with statistics such as success frequency and average speedup, allowing the library to adapt to new workloads and hardware characteristics.

\paragraph{Structure of Experience Hints}
Each experience is stored as a structured hint entry to support effective retrieval and guidance. 
A hint includes trigger conditions (e.g., error types or bottlenecks), context descriptors, optimization suggestions, and confidence statistics. 
This structure enables the framework to retrieve relevant hints based on diagnostic signals and prioritize those with consistent effectiveness.

Overall, the Experience Library acts as a dynamic knowledge base that continuously improves evolutionary guidance and enhances the efficiency and robustness of kernel optimization.

\section{Evaluation}
\begin{table*}[t]
\scriptsize
\centering

{
\begin{tabular}{lcccccccc}
\toprule
\multirow{1}{*}{\textbf{Model}}  &
\multicolumn{4}{c}{\textbf{Level 1}} &
\multicolumn{4}{c}{\textbf{Level 2}} \\
\midrule
 & \textbf{Correctness} & \textbf{Fast$_1$}& \textbf{Avg\_speedup} & \textbf{Geomean}
 & \textbf{Correctness} & \textbf{Fast$_1$}& \textbf{Avg\_speedup} & \textbf{Geomean}\\
\midrule
DeepSeek-V3 & 52\% & 6 & 0.71 & 0.56 & 23\% & 17 & 0.26 & 1.09\\

KernelLLM & 55\% & 21 & 0.49 & 0.80 & 32\% & 31  & 0.45 & 1.38\\
KernelLLM (w/o cheating) & 19\% & 3 & 0.13 & 0.51 & 28\% & 27 & 0.39 & 1.35 \\

AutoTriton & 84\% & 21 & 1.09 & 0.66 & 94\%& 58 & 0.99 &1.01\\
AutoTriton (w/o cheating) & 76\% & 11 & 0.98 & 0.56  & 92\%& 50 & 0.95  & 0.98\\

\midrule
\multicolumn{9}{c}{\textit{Kernel Foundry}} \\
\midrule
\text{Optimizer: ChatGPT-5.4}& 96\% & 31  & 1.14& 0.70  & 98\%&61 & 2.46 &1.01\\
\text{Optimizer: Qwen-3.5}& 91\% & 18  & 0.87 & 0.31  & 100\%&64 & 1.06&0.94\\
\text{Optimizer: Claude-Sonnet-4.5}& 99\% & 21  & 1.22 & 0.39  & 100\%&62 & 1.03&  0.96\\
\text{Optimizer: DeepSeek-V3} &  & &  & & & & & \\
\ \ \textit{No Init.}&  55\% & 10 & 0.81 & 0.54  &40\% & 26 & 0.43 & 1.05\\
\ \ KernelLLM Init. & 60\%  & 16  & 0.91 & 0.72  & 45\% & 38 & 0.56 & 1.22\\
\ \ AutoTriton Init. & 86\% &  17 & 1.30 & 0.37 & 97\% & 71 & 1.08 & 1.17\\
\ \ + Diagnosis& 95\% &  35 & 1.39 & 0.85 & 98\% & 86 & 1.17 & 1.15\\
 \ \ + Diagnosis + Experience Library& 98\% &  41 & 1.47 & 0.91 & 99\% & 90 & 2.85 & 1.31\\
\bottomrule
\end{tabular}

\caption{\textbf{Results on KernelBench.}
We report correctness, fast$_1$, average speedup and geomean for direct LLM-based code generation baselines and Kernel Foundry.
Rows marked as ``w/o cheating'' report performance after removing kernels identified as cheating, highlighting the impact of shortcut implementations on apparent performance.
For Kernel Foundry, we report results under different optimizer configurations.
}
\label{tab:level1}
}
\end{table*}




\subsection{Evaluation Setup}

\paragraph{Hardware and Environment.}
Experiments are conducted on an NVIDIA RTX~5090 with Python~3.10. 

\paragraph{Benchmark.}
We evaluate on KernelBench, a standardized benchmark for GPU kernel generation and optimization, including both Level~1 and Level~2 tasks spanning element-wise, reduction, matrix multiplication, and convolution operators.

\paragraph{Metrics.}
We report four metrics: \emph{Correctness} (numerical equivalence to PyTorch), \emph{Fast$_1$} (number of tasks with speedup $>1\times$), \emph{Avg\_speedup} (arithmetic mean speedup), and \emph{Geomean} (geometric mean speedup computed over correct kernels only, for robustness to outliers).

\subsection{Compared Methods}

We compare our approach with representative LLM-based Triton code generation systems, including general-purpose models (e.g., DeepSeek-V3~\cite{liu2024deepseek}), specialized kernel generation models (e.g., KernelLLM), and automated Triton translators (e.g., AutoTriton). These baselines cover the dominant classes of direct code-generation approaches without evolutionary refinement.

All methods are evaluated under a unified protocol. For each task, we sample each model 10 times to account for stochasticity, and report the best valid result (best-of-10) after correctness and cheating validation.

For Kernel Foundry, we use AutoTriton in initialization by default. We use DeepSeek-V3~\cite{liu2024deepseek}, Claude-Sonnet-4.5~\cite{anthropic_claude_sonnet_4_5}, GPT-5.4~\cite{openai_gpt5_4} and Qwen-3.5~\cite{qwen3.5} as optimizers, each running 30 evolutionary iterations per task. Results on KernelBench Level~1 and Level~2 are summarized in Table~\ref{tab:level1}.

\subsection{KernelBench Performance}
\begin{table*}[t]
\centering
\begin{tabular}{lccccc}
\toprule
\textbf{L1 Category} & \textbf{\#Cases} & \textbf{Correctness}   & \textbf{Fast$_1$} & \textbf{Avg\_speedup} & \textbf{Geomean}\\
\midrule
Unary Elementwise (A) & 13& 13 & 1 & 0.83 & 0.59 \\
Matmul / GEMM (E) & 18 & 17  & 4 & 3.42 & 1.18\\
Convolution (C) & 34& 33  & 22 & 1.00 & 0.97\\
Reduction / Norm / Loss (D) & 35& 35   & 14& 1.20 &0.88 \\
\midrule
\textbf{L2 Category} &  &   &  & \\
\midrule
Conv-centered & 14& 14 & 13 & 1.33  & 1.29\\
Matmul-centered & 12 & 12  & 11 & 1.16 & 1.14 \\
Mixed reduction & 43& 43  & 39 & 1.23 & 1.20\\
Reduction-dominated & 31& 30   & 27& 6.56 & 1.58\\
\bottomrule
\end{tabular}
\caption{\textbf{Category-level performance of Kernel Foundry using DeepSeek-V3 as the optimizer, with AutoTriton initialization, diagnosis, and the Experience Library.}}
\label{tab:level1-category}
\end{table*}


As shown in Table~\ref{tab:level1}, direct LLM generation achieves limited correctness and aggregate performance, while AutoTriton provides the strongest non-evolutionary baseline. Kernel Foundry substantially improves these solutions through iterative refinement. On Level~1, the complete configuration achieves 98\% correctness, fast$_1$ of 41, and an average speedup of 1.47$\times$  with DeepSeek-V3 as optimizer. Starting from the weaker KernelLLM initialization, evolution improves correctness from 19\% to 60\% and fast$_1$ from 3 to 16, showing that the framework can repair and optimize initially low-quality kernels. Even without expert initialization, evolution improves upon direct DeepSeek-V3 generation, although a stronger expert provides a better starting point.

The framework is also effective across different optimizer models. With AutoTriton initialization, GPT-5.4, Qwen-3.5, and Claude-Sonnet-4.5 achieve 96\%, 91\%, and 99\% correctness on Level~1, respectively. Claude-Sonnet-4.5 obtains the highest correctness and a 1.22$\times$ average speedup. These results suggest that Kernel Foundry generalizes across model backbones, although the final speedup still depends on the optimization capability of the underlying model.

The improvements are more pronounced on Level~2, which contains more complex multi-operator workloads. Starting from the AutoTriton baseline without cheating, DeepSeek-V3-based evolution improves correctness from 92\% to 97\%, fast$_1$ from 50 to 71, and average speedup from 0.95$\times$ to 1.08$\times$. Adding diagnosis further increases fast$_1$ to 86 and average speedup to 1.17$\times$. With the Experience Library, the complete framework reaches 99\% correctness, fast$_1$ of 90, a 2.85$\times$ average speedup, and a 1.31$\times$ geomean speedup. Other optimizers also achieve 98\%--100\% correctness, demonstrating that the improvements are not restricted to DeepSeek-V3.

The category-level results in Table~\ref{tab:level1-category} further reveal where the performance gains originate. On Level~1, matmul/GEMM kernels achieve a 3.42$\times$ average speedup and a 1.18$\times$ geomean speedup, whereas unary and convolution kernels offer less optimization headroom. On Level~2, all categories achieve geomean speedups above 1$\times$. Reduction-dominated pipelines obtain the largest improvement, with a 6.56$\times$ average speedup and a 1.58$\times$ geomean speedup. The difference between these two metrics indicates that several highly optimizable cases contribute disproportionately to the arithmetic mean, while the geomean above 1$\times$ still reflects broadly positive improvements.

Overall, Kernel Foundry is particularly effective for multi-operator workloads, where operator fusion, intermediate-memory elimination, and cross-operator optimization provide greater performance headroom. For Level~1, the framework achieves high correctness and increases the number of kernels outperforming PyTorch, although the speedup gains remain more heterogeneous across kernel categories.

\subsection{Ablation Study}
\begin{figure*}[t]
	\centering
	\begin{subfigure}{0.32\linewidth}
		\centering
		\includegraphics[width=0.9\linewidth]{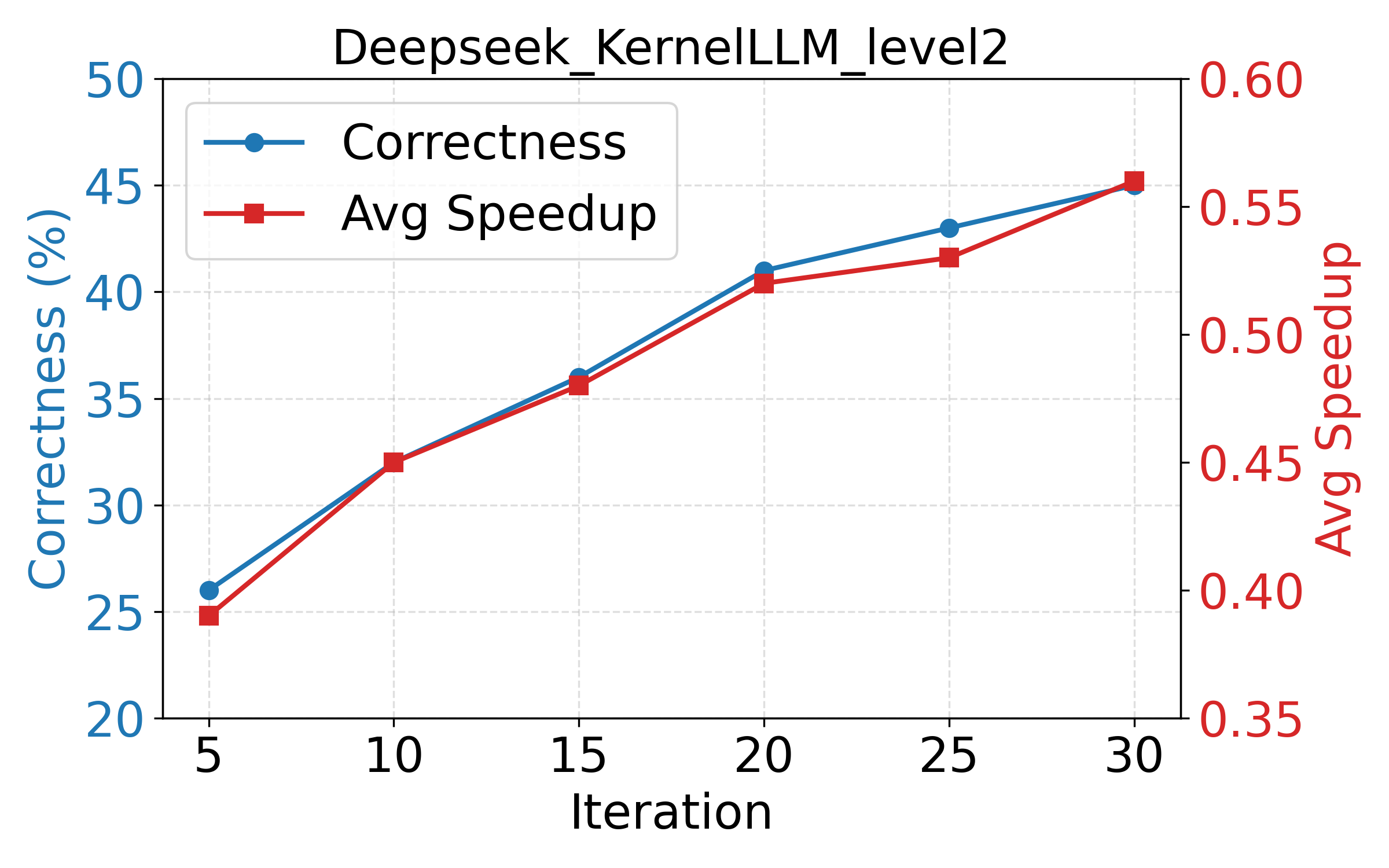}
        \vspace{-0.3cm}
		\caption{DeepSeek with KernelLLM}
        \label{fig:sub-a}
	\end{subfigure}
	\begin{subfigure}{0.32\linewidth}
		\centering
		\includegraphics[width=0.9\linewidth]{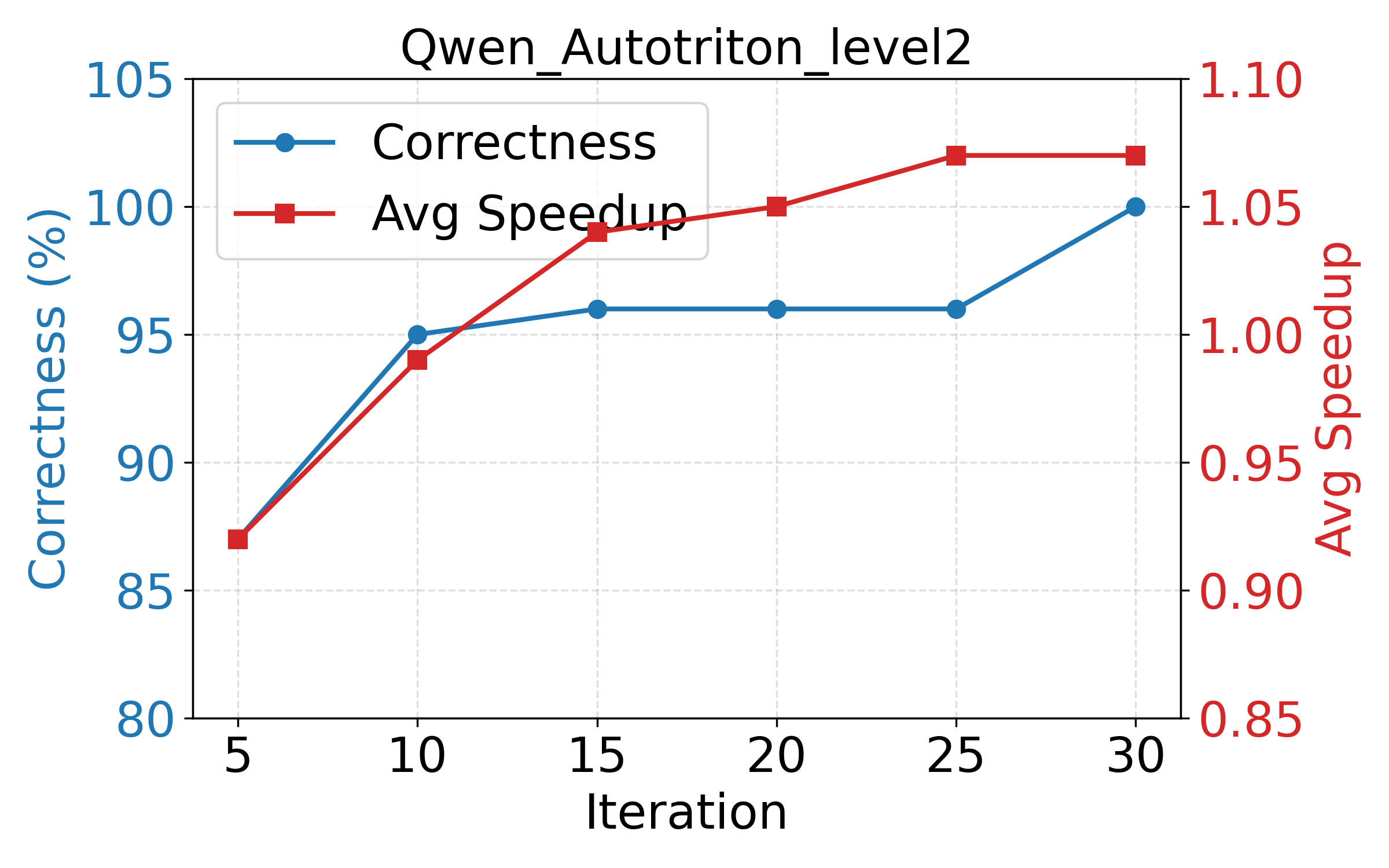}
        \vspace{-0.3cm}
		\caption{Qwen with AutoTriton}
        \label{fig:sub-a}
	\end{subfigure}
	\begin{subfigure}{0.32\linewidth}
		\centering
		\includegraphics[width=0.9\linewidth]{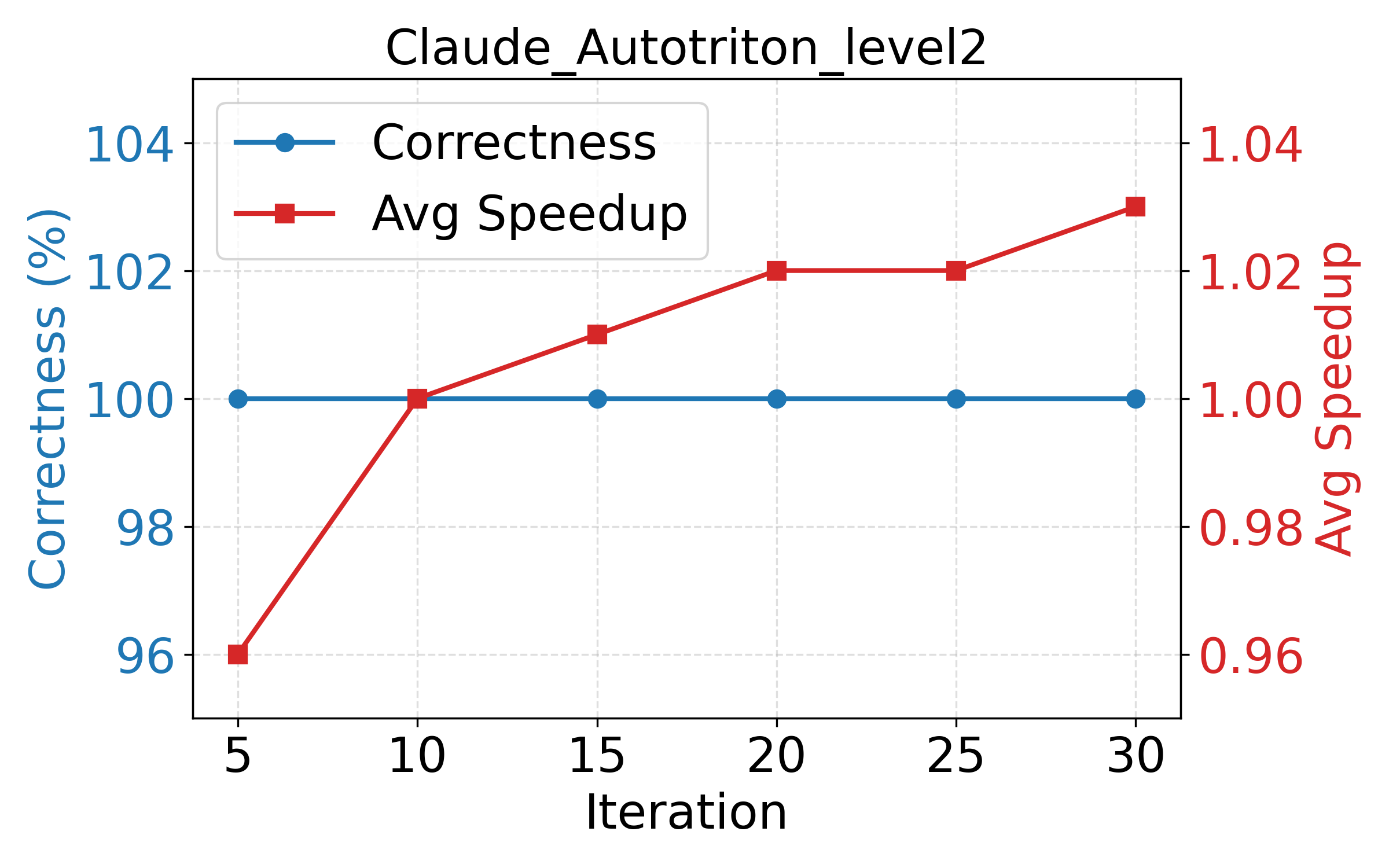}
        \vspace{-0.3cm}
		\caption{Claude with AutoTriton}
        \label{fig:sub-a}
	\end{subfigure}
    \vspace{-0.3cm}
    \caption{Correctness and average speedup as a function of evolution iterations for different settings on Level~2.}
    \label{fig:iteration_curve}
    \vspace{-0.3cm}
\end{figure*}



We conduct ablation studies on KernelBench to quantify the contribution of expert initialization, diagnosis-driven optimization, and the experience library. Unless otherwise specified, DeepSeek-V3 is used as the optimizer. The results are summarized in Table~\ref{tab:level1}.

\textbf{Evolution without expert initialization.}
We first evaluate evolution without a domain-specialized expert by directly optimizing the original LLM-generated kernels. As shown by \textit{No Init.}, evolution improves upon direct DeepSeek-V3 generation. On Level~2, correctness increases from 23\% to 40\%, while the average speedup improves from 0.26$\times$ to 0.43$\times$. On Level~1, correctness increases from 52\% to 55\%, and the average speedup improves from 0.71$\times$ to 0.81$\times$. Nevertheless, the improvements remain limited because evolution frequently starts from invalid or poorly structured kernels, making it difficult to reach high-quality regions of the optimization space.

\textbf{Impact of expert initialization.}
We next initialize the evolutionary process using kernels generated by domain-specialized expert models. KernelLLM initialization improves the Level~2 correctness from 40\% to 45\% and the average speedup from 0.43$\times$ to 0.56$\times$. Using AutoTriton provides a substantially stronger starting point, further increasing correctness to 97\%, Fast$_1$ to 71, and the average speedup to 1.08$\times$. A similar trend is observed on Level~1, where AutoTriton initialization achieves 86\% correctness and a 1.30$\times$ average speedup. These results show that expert initialization supplies useful structural priors and allows evolution to focus on meaningful kernel-level optimization rather than repeatedly repairing fundamentally invalid implementations.

\textbf{Effectiveness of diagnosis-driven optimization.}
Adding structured diagnosis on top of AutoTriton initialization consistently improves both correctness and performance. On Level~1, diagnosis increases correctness from 86\% to 95\%, Fast$_1$ from 17 to 35, and the average speedup from 1.30$\times$ to 1.39$\times$. On Level~2, it improves correctness from 97\% to 98\%, Fast$_1$ from 71 to 86, and the average speedup from 1.08$\times$ to 1.17$\times$. The particularly large gains in Fast$_1$ indicate that diagnosis does more than repair incorrect kernels: by identifying likely failure causes and performance bottlenecks, it guides mutations toward candidates that outperform the PyTorch baseline.

\textbf{Effectiveness of the experience library.}
Finally, incorporating the experience library further improves the diagnosis-driven evolutionary process. On Level~1, the full configuration reaches 98\% correctness, Fast$_1$ of 41, and an average speedup of 1.47$\times$. On Level~2, it achieves 99\% correctness, Fast$_1$ of 90, and an average speedup of 2.85$\times$, compared with 98\%, 86, and 1.17$\times$, respectively, without the library. The Level~2 geomean also improves from 1.15$\times$ to 1.31$\times$, suggesting that the performance gain is not solely caused by a single outlier. Overall, these results demonstrate that reusable optimization experience complements instance-specific diagnosis, helping the evolutionary search identify effective transformations more efficiently and achieve stronger results.

\subsection{Detailed Performance}




\textbf{Existence of cheating behaviors.}
We further analyze the impact of cheating behaviors on kernel optimization. KernelLLM exhibits a large amount of cheating, while AutoTriton shows a smaller but non-negligible portion. In most cases, cheating kernels directly fall back to the original PyTorch functions rather than implementing genuine Triton kernels. Although such code may pass correctness checks, it cannot achieve real operator acceleration. Evolution can only make superficial changes around the wrapper code. Our framework mitigates this issue by suppressing cheating behaviors, enabling effective kernel-level optimization.
We also provide a case study in the Appendix.

\textbf{Evolution iteration performance.}
Figure~\ref{fig:iteration_curve} shows how correctness and average speedup evolve with the number of evolution iterations across different settings. 
Both metrics improve steadily, with rapid gains in early iterations followed by gradual convergence. 
This trend suggests that the evolutionary process can effectively refine kernels over time while avoiding early saturation.


Overall, these results highlight the important properties of our evolution framework. Performance generally improves with additional iterations,
whereas correctness often stabilizes earlier once valid kernels
are discovered. This demonstrates that our evolutionary process is capable of continuously refining kernels and converging toward higher-quality GPU implementations over time. We also provide a case study in the Appendix.










\section{Conclusion}
We propose a diagnosis-driven evolutionary framework for automatic generation and optimization of high-performance Triton kernels. By integrating expert-guided initialization with a multi-island evolutionary search, the framework treats kernel synthesis as an iterative optimization process rather than one-shot code generation. Structured diagnosis converts evaluation outcomes into actionable feedback, and a centralized experience library accumulates reusable optimization knowledge to guide subsequent evolution. Experiments on KernelBench show that our approach consistently improves both correctness and performance over direct LLM-based generation and existing automated kernel systems such as KernelLLM and AutoTriton. The framework is robust across different optimizers and achieves the strongest results.

\bibliographystyle{plain}
\bibliography{references}

\appendix


\newpage
\appendix
\section{System Prompt for Triton Kernel Evolution}
\label{app:evaoluation-system-prompt}

\begin{tcolorbox}[title={System Prompt: Evolver},breakable,enhanced]
You are an \textbf{Evolver} specialized in generating \textbf{custom Triton kernels} to replace PyTorch operators for performance speedups.

\medskip
\textbf{Your Mission}
\begin{itemize}
  \item Analyze the provided model architecture and identify opportunities to replace PyTorch operators with Triton kernels.
  \item You have full autonomy in selecting which operators to rewrite.
  \item You may keep some operators unchanged if replacing them does not provide meaningful benefits.
\end{itemize}

\medskip
\textbf{Allowed Optimization Strategies}
\begin{enumerate}
  \item \textbf{Direct Operator Replacement}: Implement Triton kernels that faithfully reproduce the functionality of existing PyTorch operators.
  \item \textbf{Operator Fusion}: Combine multiple sequential operators into a single Triton kernel (e.g., \texttt{matmul + relu}, \texttt{layernorm + GELU}, \texttt{softmax + dropout}).
  \item \textbf{Algorithmic Optimization}: Modify the computation strategy to improve performance (e.g., online softmax, reduced precision, layout transformation).
  \item \textbf{Multi-Operator Rewrite}: Replace multiple operators in a single iteration if beneficial.
\end{enumerate}

\medskip
\textbf{Evolution Workflow}
\begin{itemize}
  \item You will be provided with:
  \begin{itemize}
    \item Historical evolution code
    \item Current evolved code
    \item Performance metrics (speedup, runtimes, correctness and so on)
    \item Hints
  \end{itemize}
  \item Based on this context, your goal is to generate the \textbf{next evolved version} of the code.
\end{itemize}

\medskip
\textbf{You are only limited by your imagination.}

\medskip
\textbf{Example Architecture}
\begin{tcolorbox}[colback=white, colframe=black!20]
\texttt{\{example\_arch\_src\}}
\end{tcolorbox}

\medskip
\textbf{Optimized Version with Triton Kernels}
\begin{tcolorbox}[colback=white, colframe=black!20]
\texttt{\{example\_new\_arch\_src\}}
\end{tcolorbox}

\medskip
\textbf{Output Requirements}
\begin{enumerate}
  \item Optimize the architecture named \texttt{Model} with custom Triton operators while preserving full functional equivalence:
\begin{tcolorbox}[colback=white, colframe=black!20]
\texttt{\{initial\_param\_code\}}
\end{tcolorbox}

  \item Generate a single, complete, and syntactically correct Python code block named \texttt{ModelNew}. Output only the new model code, with no additional text and no testing code.
  \item The core logic must be implemented in a Triton kernel decorated with \texttt{@triton.jit}.
  \item Always include the following imports:
\begin{tcolorbox}[colback=white, colframe=black!20]
\texttt{import torch\\
import triton\\
import triton.language as tl}
\end{tcolorbox}

  \item Define each function with exactly the required signature. Do not change parameter names, counts, or order. Use PyTorch tensor type hints and use \texttt{tl.constexpr} only for compile-time constants.
  \item Carefully manage data types and use Triton operations (\texttt{tl.load}, \texttt{tl.store}, \texttt{tl.dot}, \texttt{tl.arange}, masks, and \texttt{tl.math}) correctly.
  \item Assume Triton version 3.1.0 or later.
\end{enumerate}

\medskip
\textbf{Final Verification}
\begin{enumerate}
  \item All function signatures exactly match the required definitions.
  \item All function calls match their definitions.
  \item No undefined functions are called.
  \item No required parameters are missing.
\end{enumerate}
\end{tcolorbox}

\section{Expert Recommendations for Triton Kernel Optimization}
\label{app:expert-doc}
\begin{tcolorbox}[colback=gray!6, colframe=black!50, boxrule=0.5pt, arc=2pt, left=5pt, right=5pt, top=5pt, bottom=5pt, breakable]
\textbf{Curated Expert Recommendations:}

\textbf{Correctness}
\begin{enumerate}
\item Ensure numerical stability by normalizing data before exponentiation to prevent overflow, and use float32 for intermediate computations along with high-precision accumulators to reduce errors in accumulation.
\item Follow API usage constraints strictly: avoid \texttt{return}, \texttt{break}, or \texttt{continue} in kernels and use masks instead; avoid lambda expressions and chained boolean operations, replacing them with inline functions or stepwise mask computations; avoid direct tensor indexing and use \texttt{tl.load} and \texttt{tl.store}.
\item Use \texttt{tl.constexpr} only for compile-time kernel parameters, such as block sizes or flags that control kernel structure, and never on the host side or in kernel launch functions.

\item Maintain a systematic debugging checklist: verify all loads/stores have masks or boundary checks, strides are correct, array indexing does not exceed bounds, control flow uses masks appropriately, atomic operations are correctly applied for concurrent writes, and performance-related configurations (BLOCK\_SIZE, memory access, grid size) are appropriate.

\item Follow development best practices: write descriptive variable names, include sufficient comments explaining computation logic, and keep kernel functions concise and clear.

\item For convolution kernels, ensure that PyTorch random weights are replicated on the Triton host side using the same module and device as in PyTorch. Fix random seeds before kernel execution to maintain reproducibility, and ensure parameter names and module calls match the original PyTorch module.
\item When debugging kernels, check for grid and program ID mismatches, e.g., launching a 1D grid while the kernel expects 2D program IDs, and ensure program IDs are correctly mapped inside the kernel to avoid runtime errors.

\item Introduce controlled approximation techniques where exact precision is unnecessary: reduce intermediate precision selectively, enable early termination for iterative computations, and apply approximate or statistically unbiased accumulation to trade minimal accuracy loss for significant performance gains.
\item Handle precision explicitly and consistently: mix precisions only where numerically safe, avoid dynamic scaling inside kernels, apply saturation or clamping logic explicitly when required, and prefer deterministic rounding unless stochastic rounding provides measurable benefits.

\end{enumerate}
\textbf{Memory-bound}
\begin{enumerate}
\item For memory access optimization, maintain contiguous and local memory access patterns. Use \texttt{tl.make\_block\_ptr} with \texttt{boundary\_check} for 2D data and carefully design stride parameters to prevent performance degradation.

\item Pack data explicitly to improve vectorization and memory coalescing: reorganize inputs into structure-of-arrays (SoA) layouts, apply sub-tile packing for irregular shapes, and handle diagonal or sparse-like access patterns via pre-packed contiguous buffers.

\item Employ flexible tiling strategies: dynamically adjust tile sizes based on tensor aspect ratios, use rectangular tiles for asymmetric dimensions, and apply hierarchical tiling (register-level, shared-memory-level, global-memory-level) to maximize locality while maintaining occupancy.

\item Utilize software-managed prefetching by staging future data accesses across pipeline stages. Tune prefetch distance to balance latency hiding and cache pollution, and differentiate between temporal reuse (keep in cache) and streaming accesses (avoid cache thrashing).
\item Design parallelization schemes that minimize synchronization: decompose work recursively when beneficial, structure kernels to avoid global barriers, and rely on implicit program independence rather than explicit coordination whenever possible.
\item Optimize cache utilization by enforcing cache-line-aligned accesses, batching writes to enable write-combining, and applying sliding-window or cache-oblivious access patterns to sustain reuse across successive tiles.
\item Minimize memory access overhead by reducing pointer arithmetic, selecting stride-minimizing layouts, avoiding redundant transpositions, and choosing blocking factors that align with L1/L2 cache capacities.
\end{enumerate}
\textbf{Instruction-bound}
\begin{enumerate}
\item Choose block sizes as powers of two (e.g., 256, 512, 1024) and tune them to balance parallelism and resource usage. Avoid sizes that are excessively small or large, as they can reduce performance or limit concurrency.
\item Align kernel designs with hardware execution characteristics by explicitly unrolling compute-heavy loops when register pressure allows, and interleave arithmetic instructions with memory operations to hide global memory latency. Avoid excessive unrolling that may cause register spilling or reduce occupancy.
\item Exploit mixed-precision computation safely by promoting accumulators to higher precision (e.g., FP32 accumulation for FP16 inputs) while keeping inputs and outputs in lower precision. Fuse type conversions into load/store paths to avoid standalone cast operations and unnecessary kernel launches.
\item Favor instruction selections that map efficiently to GPU hardware: replace branches with mask-based arithmetic, maximize fused multiply-add (FMA) usage, leverage native FP16/BF16 operations when supported, and avoid instructions with high latency or low throughput.
\item Map SIMD-style parallelism onto Triton abstractions by expressing vectorized computation through block-level operations. Structure kernels to naturally pipeline FMA-heavy instruction streams and maximize instruction-level parallelism within each program instance.
\end{enumerate}
\textbf{Latency-bound}
\begin{enumerate}
\item Decompose complex operators into multiple simpler kernels when possible. Avoid overly complex kernels that are difficult to tune and debug.
\item Maximize performance by dynamically exploring key parameters such as BLOCK\_SIZE, num\_stages, and num\_warps, experimenting with alternative algorithmic implementations (e.g., naive, online, fused softmax), optimizing memory access patterns and numerical stability, and evaluating all feasible operator fusion strategies while respecting hardware resource constraints.
\item Autotune primary kernel fields systematically: choose appropriate tile sizes for GEMM or tensor contractions (BLOCK\_M, BLOCK\_N, BLOCK\_K) to balance compute density and cache locality; select pipeline depth (num\_stages) according to the number of fused GEMMs; and choose the number of warps per block (num\_warps, typically 1–16) to balance utilization, register usage, and shared memory consumption.

\item Replace control-flow-heavy logic with predicated execution whenever feasible. Convert conditional branches into mask-based arithmetic or masked loads/stores to reduce divergence and improve warp-level efficiency.

\item Select algorithms adaptively based on input size and shape: prefer direct or outer-product formulations for small or skinny tensors, switch to tiled inner-product or block GEMM-style implementations for large workloads, and introduce size-based thresholds to avoid inefficient recursive or asymptotically optimal algorithms in practice.

\item Schedule instructions to minimize critical paths: separate dependent load–compute–store chains across stages, prefetch data early using staged pipelines (\texttt{num\_stages}), and balance arithmetic intensity across warps to avoid execution port underutilization.

\item Optimize reduction patterns by using hierarchical, multi-stage reductions: perform partial reductions within registers or shared memory, cache intermediate results when reused, and avoid atomic operations unless inter-program synchronization is unavoidable.

\item Provide specialized kernel variants for common corner cases, such as very small tensors, power-of-two dimensions, or strictly contiguous layouts. Dispatch these fast paths from the host to avoid penalizing the general kernel with extra conditionals.

\item Construct software pipelines that explicitly overlap loads, computation, and stores: break long dependency chains, minimize pipeline bubbles by balancing instruction mix, and tune \texttt{num\_stages} and \texttt{num\_warps} jointly to maximize sustained throughput on the target GPU.

\item Balance workload assignment across Triton programs by partitioning work according to computational intensity rather than raw element count. Prefer fine-grained program decomposition for irregular workloads, and prioritize critical-path computations to avoid stragglers that limit overall kernel completion time.

\end{enumerate}
\end{tcolorbox}

\section{Case Study}
\subsection{Cheating Case of AutoTriton}
\label{app:cheating}
\begin{codebox}
\begin{lstlisting}[language=Python, basicstyle=\ttfamily\small]
import torch
import torch.nn as nn
import triton
import triton.language as tl

@triton.jit
def triton_softmax_kernel(
    x_ptrs,
    out_ptrs,
    xnumel,
    rnumel,
    cnumel,
    XBLOCK: tl.constexpr,
):
    pid = tl.program_id(axis=0)
    batch_idx = pid // cnumel
    spatial_idx = pid % cnumel
    base = batch_idx * (rnumel * cnumel) + spatial_idx

    max_value = float('-inf')
    for c in range(0, rnumel):
        offset = base + c * cnumel
        val = tl.load(x_ptrs + offset)
        max_value = tl.maximum(max_value, val)

    sum_value = 0.0
    for c in range(0, rnumel):
        offset = base + c * cnumel
        val = tl.load(x_ptrs + offset)
        sum_value += tl.exp(val - max_value)

    for c in range(0, rnumel):
        offset = base + c * cnumel
        val = tl.load(x_ptrs + offset)
        softmax_val = tl.exp(val - max_value) / sum_value
        tl.store(out_ptrs + offset, softmax_val)

class ModelNew(nn.Module):
    """
    Model that performs a 3D convolution, applies Softmax (via Triton), and performs two max pooling operations.
    """
    def __init__(self, in_channels, out_channels, kernel_size, pool_kernel_size):
        super(ModelNew, self).__init__()
        self.conv = nn.Conv3d(in_channels, out_channels, kernel_size)
        self.pool1 = nn.MaxPool3d(pool_kernel_size)
        self.pool2 = nn.MaxPool3d(pool_kernel_size)

    def forward(self, x):
        """
        Args:
            x: Input tensor of shape (batch_size, in_channels, depth, height, width)
        Returns:
            Output tensor of shape (batch_size, out_channels, depth', height', width') where depth', height', width' are the dimensions after pooling.
        """
        x = self.conv(x)
        shape = x.shape
        c = shape[1]
        spatial_size = shape[2] * shape[3] * shape[4]
        x_flat = x.reshape(shape[0], c, spatial_size)
        output = torch.empty_like(x_flat)
        num_programs = shape[0] * spatial_size
        triton_softmax_kernel[(num_programs,)](x_flat, output, x_flat.numel(), c, spatial_size, XBLOCK=1024)
        x = output.view(shape)
        x = self.pool1(x)
        x = self.pool2(x)
        return x
\end{lstlisting}
\end{codebox}

\textbf{Level2 Task6:}
(\textbf{6\_Conv3d\_Softmax\_MaxPool\_MaxPool}).
To better understand the limitations of AutoTriton on complex operator fusion tasks, we conduct a case study on Level2 Task6, which consists of a Conv3D--Softmax--MaxPool--MaxPool pipeline. Although AutoTriton reports a comparable performance on this task, a closer inspection reveals that the generated kernel only rewrites the \texttt{Softmax} operator, while the remaining operators are directly invoked via PyTorch library functions. The full code is provided above.

\vspace{0.5em}
\noindent\textbf{Partial Code Snippet.}
The following excerpt illustrates the key issue. While a custom Triton kernel is used for \texttt{Softmax}, both \texttt{Conv3d} and \texttt{MaxPool} operations fall back to PyTorch implementations:
\begin{codebox}
\begin{lstlisting}[language=Python, basicstyle=\ttfamily\small]
    def __init__(self, in_channels, out_channels, kernel_size, pool_kernel_size):
        super(ModelNew, self).__init__()
        self.conv = nn.Conv3d(in_channels, out_channels, kernel_size)
        self.pool1 = nn.MaxPool3d(pool_kernel_size)
        self.pool2 = nn.MaxPool3d(pool_kernel_size)
    def forward(self, x):
        x = self.conv(x)
        ...
        triton_softmax_kernel[(num_programs,)](x_flat, output, x_flat.numel(), c, spatial_size, XBLOCK=1024)
        x = output.view(shape)
        x = self.pool1(x)
        x = self.pool2(x)
        return x
\end{lstlisting}
\end{codebox}
\vspace{0.5em}
\noindent\textbf{Analysis.}
This design constitutes a form of \emph{cheating} behavior: although the overall program executes correctly and may exhibit limited speedup, the majority of computation remains encapsulated in opaque PyTorch library calls. As a result, these operators cannot be further analyzed, transformed, or optimized by downstream kernel evolution or mutation stages. In particular, critical optimization opportunities such as operator fusion, memory layout reorganization, and cross-operator scheduling are completely blocked.

Moreover, since performance-critical components are hidden behind black-box library calls, the reported speedup does not reflect genuine kernel-level optimization. This behavior artificially inflates performance metrics while significantly constraining the available optimization space, making the resulting code unsuitable for iterative evolution or fine-grained performance diagnosis.

In contrast, our framework enforces end-to-end kernel transparency by requiring all major operators to be explicitly implemented at the kernel level. This design choice ensures that every component remains optimizable throughout evolution, enabling meaningful performance improvements rather than superficial gains.
\subsection{Evolution of the Code}
\label{app:evolution}
\textbf{Level2 Task95: 
(\textbf{95\_Matmul\_Add\_Swish\_Tanh\_GELU\_Hardtanh})}.
For this task, the evolution mainly improves the \emph{fused activation} following the matmul (Linear) layer. At iteration 15, our framework successfully writes the correct code. Then the speedup increases from \textbf{1.04 $\rightarrow$ 1.27 $\rightarrow$ 1.29 $\rightarrow$ 1.31}, and the gains come from progressively reducing kernel overhead and tightening the fused math/dataflow. The code is provided below.
\paragraph{Iteration 20 Code (speedup $\approx$ 1.27): establishing a full fused activation kernel.}
The first strong candidate introduces a dedicated Triton kernel that \emph{fully fuses} the post-matmul chain:
\texttt{Add (bias-like)} $\rightarrow$ \texttt{Swish} $\rightarrow$ \texttt{Tanh} $\rightarrow$ \texttt{GELU (erf)} $\rightarrow$ \texttt{Hardtanh}.
It linearizes the output tensor, uses \texttt{offsets \% out\_features} to map each element to its column-wise addend, and writes the final result in one pass, eliminating multiple PyTorch launches and intermediate tensors.
This version also introduces autotuning over \texttt{BLOCK\_SIZE} and warp counts to pick a reasonable configuration across shapes, and uses a clean wrapper to handle contiguity and output allocation. :contentReference[oaicite:0]{index=0}

\paragraph{Iteration 25 Code (speedup $\approx$ 1.29): reducing launch and allocation overhead (in-place + fixed tile).}
The next iteration keeps the same fused math, but removes the autotune/search machinery and the separate output allocation by writing \emph{in-place} (i.e., \texttt{output\_ptr} is the same as \texttt{x\_ptr}). It also fixes \texttt{BLOCK\_SIZE=1024} directly at the call site, which avoids autotune overhead and simplifies dispatch.
Practically, this stage trades some portability for lower constant overhead, which is beneficial when the fused activation is memory/latency sensitive relative to matmul output size. :contentReference[oaicite:1]{index=1} :contentReference[oaicite:2]{index=2}

\paragraph{Iteration 30 Code  (speedup $\approx$ 1.31): algebraic and instruction-level tightening inside the fusion.}
The final improvement is dominated by micro-optimizations inside the fused function:
(i) reusing \texttt{sigmoid(x)} instead of recomputing it implicitly,
(ii) computing \texttt{exp(2*swish)} once and reusing it for the tanh transform,
(iii) hoisting constants and intermediate products (e.g., \texttt{0.7071...}) into named temporaries, and
(iv) structuring the computation to reduce redundant conversions and temporaries before \texttt{erf} and clamping.
These changes reduce instruction count and register pressure in the hot loop while preserving the one-pass fused dataflow, yielding the last incremental gain from 1.29 to 1.31. :contentReference[oaicite:3]{index=3}

\paragraph{Summary.}
Overall, the evolution first achieves \emph{end-to-end fusion} of the activation chain (major gain), then improves \emph{runtime overhead} via in-place execution and simplified dispatch, and finally extracts extra performance through \emph{math reuse and instruction reduction} within the fused kernel.

\textbf{Iteration 15 Code}
\begin{codebox}
    \begin{lstlisting}[language=Python, basicstyle=\ttfamily\small]

def fused_activation_kernel(
    x_ptr,
    add_value_ptr,
    output_ptr,
    numel: tl.constexpr,
    out_features: tl.constexpr,
    BLOCK_SIZE: tl.constexpr,
):
    pid = tl.program_id(axis=0)
    block_start = pid * BLOCK_SIZE
    offsets = block_start + tl.arange(0, BLOCK_SIZE)
    mask = offsets < numel

    x = tl.load(x_ptr + offsets, mask=mask)
    col_idx = offsets % out_features
    add_value = tl.load(add_value_ptr + col_idx, mask=mask)

    x = x + add_value
    
    swish = tl.sigmoid(x) * x
    
    tanh_out = (tl.exp(2.0 * swish) - 1.0) / (tl.exp(2.0 * swish) + 1.0)
    
    gelu = tanh_out * 0.5 * (1.0 + tl.erf(tanh_out * 0.7071067811865475))
    
    hardtanh = tl.minimum(tl.maximum(gelu, -1.0), 1.0)
    
    tl.store(output_ptr + offsets, hardtanh, mask=mask)


def fused_activation_triton(x: torch.Tensor, add_value: torch.Tensor) -> torch.Tensor:
    if not x.is_contiguous():
        x = x.contiguous()
    
    output = torch.empty_like(x)
    numel = x.numel()
    out_features = add_value.shape[0]
    
    grid = lambda meta: (triton.cdiv(numel, meta['BLOCK_SIZE']),)
    
    fused_activation_kernel[grid](
        x, add_value, output,
        numel, out_features
    )
    
    return output
    \end{lstlisting}
\end{codebox}
\textbf{Iteration 20 Code}
\begin{codebox}
    \begin{lstlisting}[language=Python, basicstyle=\ttfamily\small]

@triton.jit
def fused_activation_kernel(
    x_ptr,
    add_value_ptr,
    output_ptr,
    n_elements,
    out_features,
    BLOCK_SIZE: tl.constexpr,
):
    pid = tl.program_id(axis=0)
    block_start = pid * BLOCK_SIZE
    offsets = block_start + tl.arange(0, BLOCK_SIZE)
    mask = offsets < n_elements
    
    x = tl.load(x_ptr + offsets, mask=mask)
    col_idx = offsets % out_features
    add_val = tl.load(add_value_ptr + col_idx, mask=mask)
    
    x = x + add_val
    swish = tl.sigmoid(x) * x
    
    tanh_val = (tl.exp(2 * swish) - 1) / (tl.exp(2 * swish) + 1)
    
    gelu = tanh_val * 0.5 * (1.0 + tl.erf(tanh_val * 0.7071067811865475))
    hardtanh = tl.where(gelu > 1.0, 1.0, tl.where(gelu < -1.0, -1.0, gelu))
    
    tl.store(output_ptr + offsets, hardtanh, mask=mask)
    \end{lstlisting}
\end{codebox}
\textbf{Iteration 25 Code}
\begin{codebox}
    \begin{lstlisting}[language=Python, basicstyle=\ttfamily\small]

@triton.jit
def fused_activation_kernel(
    x_ptr,
    add_value_ptr,
    output_ptr,
    n_elements,
    out_features,
    BLOCK_SIZE: tl.constexpr,
):
    pid = tl.program_id(axis=0)
    block_start = pid * BLOCK_SIZE
    offsets = block_start + tl.arange(0, BLOCK_SIZE)
    mask = offsets < n_elements
    
    x = tl.load(x_ptr + offsets, mask=mask)
    col_idx = offsets % out_features
    add_val = tl.load(add_value_ptr + col_idx, mask=mask)
    
    x = x + add_val
    swish = tl.sigmoid(x) * x
    
    tanh_val = (tl.exp(2 * swish) - 1) / (tl.exp(2 * swish) + 1)
    
    gelu = tanh_val * 0.5 * (1.0 + tl.erf(tanh_val * 0.7071067811865475))
    hardtanh = tl.where(gelu > 1.0, 1.0, tl.where(gelu < -1.0, -1.0, gelu))
    
    tl.store(output_ptr + offsets, hardtanh, mask=mask)
    \end{lstlisting}
\end{codebox}
\textbf{Iteration 30 Code}
\begin{codebox}
    \begin{lstlisting}[language=Python, basicstyle=\ttfamily\small]

@triton.jit
def fused_activation_kernel(
    x_ptr,
    add_value_ptr,
    output_ptr,
    n_elements,
    out_features,
    BLOCK_SIZE: tl.constexpr,
):
    pid = tl.program_id(axis=0)
    block_start = pid * BLOCK_SIZE
    offsets = block_start + tl.arange(0, BLOCK_SIZE)
    mask = offsets < n_elements
    
    x = tl.load(x_ptr + offsets, mask=mask)
    col_idx = offsets % out_features
    add_val = tl.load(add_value_ptr + col_idx, mask=mask)
    
    x = x + add_val
    sigmoid_x = tl.sigmoid(x)
    swish = sigmoid_x * x
    
    exp_2swish = tl.exp(2.0 * swish)
    tanh_val = (exp_2swish - 1.0) / (exp_2swish + 1.0)
    
    gelu_input = tanh_val * 0.7071067811865475
    erf_val = tl.erf(gelu_input)
    gelu = tanh_val * 0.5 * (1.0 + erf_val)
    
    hardtanh = tl.where(gelu > 1.0, 1.0, tl.where(gelu < -1.0, -1.0, gelu))
    
    tl.store(output_ptr + offsets, hardtanh, mask=mask)
    \end{lstlisting}
\end{codebox}

\section{Limitations} While our framework demonstrates strong performance on KernelBench, several limitations remain. 

\paragraph{Computational Cost.}
Our framework incurs non-trivial computational overhead due to iterative kernel generation, compilation, and execution. 
For a representative setting (GPT-5.4 on KernelBench Level~2), a full run takes $\sim$18.5 hours, consumes $\sim$24M tokens, and costs about \$100. 
While this overhead is higher than one-shot generation methods, it enables substantially improved correctness and performance, and can be partially amortized as the experience library accumulates reusable optimization knowledge.

\textbf{Dependence on initialization quality.} The effectiveness of evolution is influenced by the quality of initial kernels. Poor initial candidates may slow convergence or restrict the search to suboptimal regions of the optimization space. While our expert-guided initialization alleviates this issue, the framework may still struggle when no reasonable starting point is available. 

\textbf{Sensitivity to evaluation signals.} The diagnosis process relies on lightweight execution signals and runtime statistics. In cases where these signals are coarse or noisy, the inferred performance bottlenecks may be inaccurate, which can affect the quality of generated optimization hints and subsequent evolution.

\end{document}